\documentclass[preprint,12pt]{elsarticle}

\usepackage{pgfplots}
\pgfplotsset{compat=1.5}
\pgfplotsset{every axis/.append style={thick}}
\usepackage{hyperref}
\usepackage{amsmath,amsfonts,amssymb}
\usepackage{changes}
\usepackage{enumerate}
\usepackage{booktabs}
\usepackage{algorithm}
\usepackage{algpseudocode}
\usepackage{multirow}
\usepackage{subcaption}
\usepackage{float}
\usepackage{xcolor}
\usepackage{tikz}

\usetikzlibrary{shapes.geometric, arrows}

\journal{Elsevier}

\begin{document}

\begin{frontmatter}



\title{CSF: Fixed-outline Floorplanning Based on the Conjugate Subgradient Algorithm Assisted by Q-Learning\tnoteref{grant}}
\tnotetext[grant]{This work is supported in part by the Fundamental Research Funds for the Central University (No. 104972024KFYjc0055), the National Natural Science Foundation of China (No. 92373102,12161039), the Natural Science Foundation of Jiangxi Province (No. 20224BAB201011), and the Foundation of Wuhan Institute of Technology (No. K2024045).}

\author[label1]{Xinyan Meng\corref{cor2}}
\author[label2]{Huabin Cheng\corref{cor2}}
\author[label1]{Rujie Chen}
\author[label4]{Ning Xu}
\author[label1]{Yu Chen\corref{cor1}}\ead{ychen@whut.edu.cn}
\author[label3]{Wei Zhang\corref{cor1}}\ead{wzhang\_math@whu.edu.cn}

\affiliation[label1]{organization={School of Mathematics and Statistics},
            addressline={Wuhan University of Technology},
            city={Wuhan},
            postcode={430070},
            country={China}}
\affiliation[label2]{organization={Department of Basic Science},
            addressline={Wuchang Shouyi University},
            city={Wuhan},
            postcode={430064},
            country={China}}
\affiliation[label4]{organization={School of Information Engineering},
            addressline={Wuhan University of Technology},
            city={Wuhan},
            postcode={430070},
            country={China}}
\affiliation[label3]{organization={School of Mathematics and Physics},
            addressline={Wuchang Institute of Technology},
            city={Wuhan},
            postcode={430305},
            country={China}}

\cortext[cor2]{X. Meng and H. Cheng contributed equally to this research.}
\cortext[cor1]{Corresponding authors.}


\begin{abstract}
The state-of-the-art researches indicate that analytic algorithms are promising in handling complex floorplanning scenarios. However, it is challenging to generate compact floorplans with excellent wirelength optimization effect due to the local convergence of gradient-based optimization algorithms designed for constructed smooth optimization models. Accordingly, we propose to construct a nonsmooth analytic floorplanning model addressed by the conjugate subgradient algorithm (CSA), which is accelerated by a population-based scheme adaptively regulating the stepsize with the assistance of Q-learning. In this way, the proposed CSA assisted by Q-learning (CSAQ) can strike a good balance on exploration and exploitation.
Experimental results on the MCNC and GSRC benchmarks demonstrate that the proposed fixed-outline floorplanning algorithm based on CSAQ (CSF) not only address global floorplanning effectively, but also get legal floorplans more efficiently than the constraint graph-based legalization algorithm as well as its improved variants. It is also demonstrated that the CSF is competitive to the state-of-the-art  algorithms on floorplanning scenarios only containing hard modules.

\end{abstract}


\begin{highlights}
\item The nonsmooth wirelength and overlapping metrics are directly optimized by conjugate subgradient algorithm (CSA). Since the CSA searches the solution space by both the deterministic gradient-based search and the stochastic exploration, it is much more likely to get the optimal floorplan by flexibly regulating its stepsize.
    \item The CSA is implemented in a population-based search pattern equipped with a dynamical stepsize regulated based on Q-learning, which contributes to its fast convergence to the optimal floorplan.
    \item Two improved variants  are proposed for the constraint graph-based legalization algorithm. They take less time to get legal floorplans with shorter wirelength.
     \item We propose a CSA-based fixed-outline floorplanning (CSF) algorithm, where both global floorplanning and legalization are implemented by the CSA assisted by Q-learning (CSAQ). It can strike a balance between time complexity and floorplan quality, and demonstrates superior performance to the compared algorithms.
\end{highlights}

\begin{keyword}
Fixed-outline floorplanning \sep nonsmooth analytic optimization \sep conjugate subgradient algorithm \sep Q-learning \sep very large-scale integration circuit



\end{keyword}

\end{frontmatter}



\section{Introduction}
Floorplanning is a key stage in the physical design flow of very large-scale integration circuit (VLSI)~\cite{Srinivasan2023}. With the rapid development of integrated circuit manufacturing technology, the complexity of VLSI design has increased dramatically, and modern design usually focuses on fixed-outline floorplanning (FOFP) problems~\cite{Kahng2001}. Compared to layout planning problems without fixed outline, the  FOFP problem is more challenging due to the additional constraint of placing a large number of modules within a given outline~\cite{Saurabh2003}. Additionally, the usage of intellectual property (IP) core introduces hard modules in the VLSI floorplan, making it much more challenging to optimize the fixed-outline floorplan of mixed-size modules~\cite{Weng2018}. 

Floorplanning algorithms generally fall into two categories: floorplanning algorithms based on combinatorial optimization model (FA-COM) and floorplanning algorithms based on analytic optimization model (FA-AOM). Coding the relative positions of modules by combinatorial data structures, FA-COM formulates floorplanning problems as  combinatorial optimization problems addressed by heuristics and metaheuristics~\cite{2006Fast-FA,Chen2010,Zou2016,Ye2020}. Although the combinatorial codes can be naturally converted into compact floorplans complying with the non-overlapping constraint, the combinatorial explosion contributes to their poor performance on large-scale cases. A remedy for this challenge is to implement multilevel methods, in which clustering or partitioning strategies are introduced to reduce the scale of the problem~\cite{Saurabh2003}. However, the multilevel strategy typically optimizes the floorplan in a local way, which in turn makes it difficult to converge to the globally optimal floorplan of large-scale VLSI designs~\cite{Yang2010,JI2021105225}.

FA-AOM address analytical floorplanning models by continuous optimization algorithms, contributing to their low complexity and fast convergence in large-scale cases~\cite{Huang2021}.
A typical FA-AOM consists of two stages, the global floorplanning stage generating promising results, and the legalization stage that refines the positions of modules to eliminate constraint violations. To get a nice VLSI floorplan, the cooperative design of global floorplanning and legalization is extremely challenging due to the complicated characteristics of floorplanning models. The efficient global floorplanning is typically implemented by addressing a smoothed optimization model with a first-order gradient-based analytical algorithm~\cite{Li2023,Huang2023}, however, the smoothing approximation could introduce some distortion to the characteristics of the mathematical model, and the first-order analytical algorithm usually converges to a local optimal solution. Legalization can be implemented through the refinement of constraint graphs~\cite{Li2023,Huang2023} or the second-order cone programming strategy~\cite{Lin2011}, which cannot strike a good balance between time complexity and minimization of wirelength.


To address the fixed-outline floorplanning problem efficiently, this paper proposes a fixed-outline floorplanning algorithm based on the conjugate subgradient algorithm (CSA), where the Q-learning is introduced to adaptively regulate the step size of CSA. The main contributions of this paper are as follows.
\begin{itemize}
    \item The nonsmooth wirelength and overlapping metrics are directly optimized by CSA. Since the CSA searches the solution space by both the deterministic gradient-based search and the stochastic exploration, it is much more likely to get the optimal floorplan by flexibly regulating its step size.
    \item The CSA is implemented in a population-based search pattern equipped with a dynamical stepsize regulated based on Q-learning, which contributes to its fast convergence to the optimal floorplan.
    \item Two improved variants are proposed for the CG-based legalization algorithm~\cite{Li2023,Huang2023}. They take less time to get legal floorplans with shorter wirelength.
     \item We propose a CSA-based fixed-outline floorplanning (CSF) algorithm, where both global floorplanning and legalization are implemented by the CSA assisted by Q-learning (CSAQ). It can strike a balance between time complexity and floorplan quality, and demonstrates superior performance to the compared algorithms.
\end{itemize}

The remainder of this paper is organized as follows. Section \ref{sec:Re} reviews related work. Section \ref{sec:Pre} introduces some preliminaries. Then, the proposed algorithm developed for fixed-outline floorplanning problems is presented in Sections \ref{sec:Meth}. The numerical experiment is performed in Section \ref{sec:Exp} to demonstrate the competitiveness of the proposed algorithm, and finally, this paper is concluded in Section \ref{sec:Con}.

\section{Related Work}\label{sec:Re}
By formulating the floorplanning problem as an analytic optimization model, Zhan \emph{et al.}~\cite{zhan2006} proposed an algorithm framework consisting of the global floorplanning stage and the legalization stage, demonstrating performance superior to the \emph{Parquet-4} based on sequence pair and simulated annealing~\cite{Saurabh2003}.  Lin and Hung~\cite{Lin2011} modelled the global floorplanning as a convex optimization problem, where modules are transformed into circles, and a push-pull mode is proposed with consideration of wirelength. Then, a second-order cone programming is deployed to legalize the floorplan.
By incorporating the electrostatic field model proposed in \emph{eplace}~\cite{Lu2015}, a celebrated algorithm for standard cell placement of VLSI, Li \emph{et al.}~\cite{Li2023} proposed an analytic floorplanning algorithm for large-scale floorplanning cases. Then, the horizontal constraint graph (HCG) and the vertical constraint graph (VCG) were constructed to eliminate overlap of floorplanning results. Huang \emph{et al.}~\cite{Huang2023} further improved the electrostatics-based analytical method for fixed-outline floorplanning, in which  module rotation and sizing were introduced to get floorplanning results with shorter wirelength. Yu \emph{et al.}~\cite{2025Per-RMAP} developed a floorplanning algorithm based on the feasible-seeking approach, achieving better performance than the state-of-the-art floorplanning algorithms.

Since the typical wirelength and overlap metrics of analytical floorplanning models are not smooth, additional smoothing approximation must be incorporated to achieve the feasibility of gradient-based optimization algorithms. Popular smoothing approximation methods for the half-perimeter wirelength (HPWL) include the quadratic model, the logarithm-sum-exponential model, and the $L_p$-norm model, and the bell-shaped smoothing method can be adopted to smooth the overlap metric~\cite{Chang2009}. In addition, Ray \emph{et al.}~\cite{Ray2013} proposed a non-recursive approximation to simulate the max function for the non-smooth HPWL model, Chan \emph{et al.}~\cite{Chan2006} employed a Helmholtz smoothing technique to approximate the density function.  However, the additional smoothing process not only introduces extra time complexity of the FA-AOM, but also leads to its local convergence to a solution significantly different from that of the original non-smooth model. Zhu \emph{et al.}~\cite{Zhu2015} proposed to address the non-smooth analytical optimization model of cell placement by the CSA, and Sun \emph{et al.}~\cite{Sun2024} further developed a CSA-based floorplanning algorithm for scenarios without soft modules.

Recently, machine learning (ML) algorithms were introduced to design general floorplanning algorithms. The generalization ability of well-trained ML algorithms could contribute their potential applications in diverse scenarios of floorplanning, but it is challenging to get an excellent ML-based floorplanning algorithm in the case of inadequate training data. He \emph{et al.}~\cite{He2020} employed agents guided by the Q-learning algorithm, which alleviates the requirement of too much prior human knowledge in the search process. Mirhoseini \emph{et al.}~\cite{Mirhoseini2021} proposed a RL-based method for chip floorplanning, and  developed an edge-based graph convolutional neural network architecture capable of learning the transferability of chips. By merging RL with the graph convolutional network (GCN), Xu \emph{et al.}~\cite{Xu2022} proposed an end-to-end learning-based floorplanning framework named as \emph{GoodFloorplan}. Yang \emph{et al.}~\cite{Yang2024} proposed an end-to-end reinforcement learning (RL) framework to learn a policy of floorplanning, and promising results were obtained with the assistance of edge-augmented graph attention network, position-wise multi-layer perceptron, and gated self-attention mechanism.

\section{Preliminaries}\label{sec:Pre}
\subsection{The Non-smooth Analytical Optimization Model of Floorplanning}\label{sec:pre-1}
Let $\mathbf{x}=(x_1,x_2,\dots,x_n)$ and $\mathbf{\mathbf{y}}=(y_1,y_2,\dots,y_n)$ be the vectors of $x$-coordinate and $y$-coordinate of all modules, respectively. 
The FOFP aims to place modules within a fixed outline such that all of them are mutually non-overlapping, formulated as
\begin{equation} \label{prob:FOFP}
\begin{array}{rl}
   \min  &  W(\mathbf{x},\mathbf{y}),\\
   s.t.  &  \begin{cases}
       D(\mathbf{x},\mathbf{y})=0,\\
       B(\mathbf{x},\mathbf{y})=0,\\
       \mathbf{x},\mathbf{y}\in\mathbf{R}^n,
   \end{cases}
\end{array}
\end{equation}
where $W(\mathbf{x}, \mathbf{y})$, $D(\mathbf{x}, \mathbf{y})$, $B(\mathbf{x}, \mathbf{y})$ are the analytical metrics of wirelength, overlapping and boundary violation, respectively.
\paragraph{The Total Wirelength $W(  \mathbf{x},  \mathbf{y})$} The total wirelength is here taken as the sum of half-perimeter wirelength (HWPL)
	\begin{equation}\label{obj:HWPL}
		W(  \mathbf{x},  \mathbf{y})=\sum_{e\in E} (\max\limits_{v_i \in e}  x_i-\min\limits_{v_i \in e}  x_i +\max\limits_{v_i \in e}  y_i-\min\limits_{v_i \in e}  y_i)
	\end{equation}
	
	
\paragraph{The Total Overlapping Area $D(  \mathbf{x},  \mathbf{y})$} The sum of overlapping area is computed by
\begin{equation}\label{obj:SOA}
		D(  \mathbf{x},  \mathbf{y})=\sum_{i,j}O_{i,j}(  \mathbf{x}) \times O_{i,j}(  \mathbf{y})
\end{equation}
where $O_{i,j}(\mathbf{x})$ and $O_{i,j}(  \mathbf{y})$ represent the overlapping lengths of module $i$ and $j$ in the $X$-axis and $Y$-axis directions, respectively.
Denoting $\Delta_x(i,j)=|x_i-x_j|$, we know
\begin{equation}\label{ind:OLX}
O_{i,j}(  \mathbf{x})=
	\begin{cases}
		\min(\hat w_i,\hat w_j), & \mbox{if } 0\le \Delta_x(i,j) \le \frac{|\hat w_i-\hat w_j|}{2}\\
		\frac{\hat w_i-2\Delta_x(i,j)+\hat w_j}{2}, & \mbox{if } \frac{|\hat w_i-\hat w_j|}{2} < \Delta_x(i,j) \le \frac{\hat w_i+\hat w_j}{2}\\
		0, &  \mbox{if }  \frac{\hat w_i+\hat w_j}{2} < \Delta_x(i,j)
	\end{cases}
\end{equation}
 Denoting $\Delta_y(i,j)=|y_i-y_j|$, we have
\begin{equation}\label{ind:OLY}
O_{i,j}(  \mathbf{y})=
	\begin{cases}
		\min(\hat h_i,\hat h_j), & \mbox{if } 0\le \Delta_y(i,j) \le \frac{|\hat h_i-\hat h_j|}{2}\\
		\frac{\hat h_i-2\Delta_y(i,j)+\hat h_j}{2}, & \mbox{if } \frac{|\hat h_i-\hat h_j|}{2} < \Delta_y(i,j) \le \frac{\hat h_i+\hat h_j}{2}\\
		0, &  \mbox{if }  \frac{\hat h_i+\hat h_j}{2} < \Delta_y(i,j)
	\end{cases}
\end{equation}
where $\hat w_i$ and $\hat h_i$ represent the width and hight of module $i$.

\paragraph{The Boundary Violation $B(\mathbf{x}, \mathbf{y})$} For the fixed-outline floorplanning problem, the position of the $i^{th}$ module must meet the following constraints:
	\begin{equation*}
		\begin{cases}
			0 \le x_i-\hat w_i/2, \quad x_i+\hat w_i/2 \le \mathbf{W^*} \\
			0 \le y_i-\hat h_i/2, \quad y_i+\hat h_i/2 \le \mathbf{H^*}
		\end{cases}
	\end{equation*}
where $\mathbf{W^*}$ and $\mathbf{H^*}$ are the width and the height of fixed-outline, respectively. They can be manually assigned or  generated by
\begin{equation}\label{boundry}
\mathbf{W^*} = \sqrt{(1 + \gamma) \cdot A/R}, \quad \mathbf{H^*} = \sqrt{(1 + \gamma) \cdot A \cdot R},
\end{equation}
where $A$ is the area sum of modules, $R$ is the width-height ratio, and $\gamma$ is the ratio of whitespace.

Denote
{	\begin{align*}
		& b_{1,i}(  \mathbf{x})=\max (0,\hat w_i/2-x_i),b_{2,i}(  \mathbf{x})=\max (0,\hat w_i/2+x_i-\mathbf{W^*}), \\		& b_{1,i}(  \mathbf{y})=\max (0,\hat h_i/2-y_i),b_{2,i}(  \mathbf{y})=\max (0,\hat h_i/2+y_i-\mathbf{H^*}),
	\end{align*}
    }
and we define $B(  \mathbf{x},  \mathbf{y})$  as
\begin{equation}\label{obj:SWFO}
B(  \mathbf{x},  \mathbf{y})=\sum_{i=1}^{n}(b_{1,i}(  \mathbf{x})+b_{2,i}(  \mathbf{x})+b_{1,i}(  \mathbf{y})+b_{2,i}(  \mathbf{y})).
\end{equation}

For the global floorplanning stage, we first solve the unconstraint optimization problem
\begin{equation}\label{prob:gf}
\min\,\,    f_g(\mathbf{x},\mathbf{y})=\alpha W(\mathbf{x},\mathbf{y})+\lambda D(\mathbf{x},\mathbf{y})+\mu B(\mathbf{x},\mathbf{y}),
\end{equation}
where $\alpha$, $\lambda$, and $\mu$ are weight coefficients. Then, the legalization is implemented by addressing the prolem
\begin{equation} \label{prob:lg}
\min\,\, f_l(\mathbf{x},\mathbf{y})=\lambda D(\mathbf{x},\mathbf{y})+\mu B(\mathbf{x},\mathbf{y}).
\end{equation}

Note that the objective functions in \eqref{prob:gf} and \eqref{prob:lg} are not smooth and traditional gradient-based algorithms cannot be applied directly, and we address them by the CSA~\cite{Zhu2015,Sun2024}.

\subsection{The Conjugate Subgradient Algorithm}

 The pseudo code of CSA is presented in Algorithm \ref{alg:CSA}~\cite{Zhu2015very}. The iteration process of CSA starts with an initial solution $\mathbf{u}_0$ and the subgradient $\mathbf{g}_0$, and the conjugate direction is initialized as $\mathbf{d}_0=\mathbf{0}$. At generation $k$, the searching direction is determined according to the subgradient $\mathbf{g}_k$ and the conjugate direction $\mathbf{d}_k$, and the searching step size is determined by dividing a scaling factor $c$ by the norm of $\mathbf{d}_k$. When $\mathbf{u}$ lies at a non-smooth point of $f$, the computation of subgradient is performed with a random scheme. Thus, the CSA is not necessarily gradient descendant. Consequently, the adaptive regulation of step size plays a critical role and has a significant influence on its convergence performance, which can be achieved by a reinforcement learning strategy.
\begin{algorithm}[!htb]
\caption{$CSA$}
\label{alg:CSA}
\begin{algorithmic}[1] 
\small \Require \small Objective function $f(\mathbf u)$, initial solution $u_0$
\Ensure \small Optimal solution $\mathbf u^*$
\State \small Set $\mathbf g_0\in\partial f(\mathbf u_0)$, $\mathbf d_0=0$;
\For{$k = 1$ to $k_{max}$}
\State Calculate the subgradient $\mathbf g_k\in\partial f(\mathbf u_{k-1})$;
\State Calculate the \emph{Polak-Ribiere} parameter $\eta_k=\frac{\mathbf g_k ^T(\mathbf g_k-\mathbf g_{k-1})}{\| \mathbf g_{k-1}\|_2 ^2}$;
\State Calculate the conjugate direction $\mathbf d_k=-\mathbf g_k+\eta _k\mathbf d_{k-1}$;
\State Calculate the step size $\hat a_k=\frac{c}{\lVert \mathbf d_k\rVert _2}$;
\State Update the solution $\mathbf u_k=\mathbf u_{k-1}+\hat a_k\mathbf d_k$;
\State Update the optimal solution $\mathbf u^*$;
\EndFor
\end{algorithmic}
\end{algorithm}

It was proved that the CSA presented in Algorithm \ref{alg:CSA} can converge to the neighborhood of the local optimum~\cite{Zhu2015very}. However, the local optimums of \eqref{prob:gf} and \eqref{prob:lg} do not guarantee the high quality of floorplanning results. To improve the global exploration without  sacrificing the local exploitation, we propose to adaptively regulate the step size of CSA based on Q-learning.

\subsection{Q-learning}
We propose to update the scaling parameter $c$ in Algorithm \ref{alg:CSA}  by Q-learning,  a typical value-based reinforcement learning algorithm~\cite{Watkins1992}. It aims to find for each state-action pair $(s,a)$ a Q-value that represents its long-term expected return.  The Q-values record an individual's learning experience, stored in a Q-table and updated by
\begin{equation} \label{ql:update}
Q^k(s_i,a_j)=(1-\alpha _0)Q^{k-1}(s_i,a_j)+\alpha _0(R(s_i,a_j)+\gamma _0Q_{max}),
\end{equation}
where $Q^k(s_i, a_j)$ represents the Q-values of the state-action pair $(s_i, a_j)$ at steps $k$.  $\alpha_0$, $\gamma_0$ are the learning rate and the discount rate, respectively. $R(s_i,a_j)$ is the reward function. In this paper, $Q_{max}$ is confirmed as
\begin{equation} \label{ql:Qm}
Q_{max}=m_0Q^{k-1}(s_i,a_j)+\frac{1-m_0}{p-1}\sum_{l\neq j}^{p} Q_{max}^{k-1}(s_l),
\end{equation}
where $m_0$ is a preset coefficient, $Q_{max}^{k - 1}(s_l)$ is the maximum Q-value across all actions of state $l$ at iteration $k-1$.

\subsection{The Constraint Graph}
The constraint graph (CG) is a directed acyclic graph (DAG) used to represent the relative position between  modules in a layout. Modules are represented by vertexes of a CG, and arcs indicate the constraint relationships between modules. Accordingly, the horizontal constraint graph (HCG) and the vertical constraint graph (VCG) are constructed for a layout to indicate the left-right and down-up positional relationships between modules, respectively.
If the lower-left corner coordinates of module $A$ are less than those of module $B$, then the arcs of the HCG and the VCG are constructed by the method proposed as below~\cite{Li2023}.
\begin{enumerate}
\item If module $A$ and $B$ do not overlap in either horizontal or vertical direction, then add the arc $A \to B$ in the HCG and add the arc $A \uparrow B$ in the VCG.
\item If module $A$ and $B$ overlap in the vertical direction but not in the horizontal direction, then add the arc $A \to B$ in the HCG; add the arc $A \uparrow B$ in the VCG if module $A$ and $B$ overlap in the horizontal direction but not in the  vertical direction.
\item For the case that module $A$ and $B$ overlap in both horizontal and vertical directions, add the arc $A \uparrow B$ in the VCG if the overlapping length in the horizontal direction is greater than that in the vertical direction; conversely, add the arc $A \to B$ in the HCG.
\end{enumerate}

Given a set of modules to be placed, a CG pair (HCG, VCG) can be translated into a floorplan where all modules are mutually disjoint. Then, by constructing a CG pair (HCG, VCG) for the result of global floorplanning and translating it into another floorplan of modules, the overlaps between modules can be eliminated and we get a compact floorplan for the modules to be placed.

\section{The CSA-based Fixed-outline Floorplanning Algorithm}\label{sec:Meth}
As presented in Section \ref{sec:pre-1}, the global floorplanning and the legalization can be implemented by addressing \eqref{prob:gf} and \eqref{prob:lg}, respectively. Accordingly, we propose a CSA-based fixed-outline floorplanning algorithm (CSF), where the performance of CSA enhances by adaptively regulating the scaling factor $c$ of step size via Q-learning.

\subsection{The Framework of CSF}
The framework of CSF is presented as Algorithm \ref{alg:fp}. At the beginning, a solution population $Pop$ of size $p$ is initialized, where the orientations of modules are randomly generated and the latin hypercube sampling (LHS) is employed to generate the initial position vector $\mathbf{Ind}_i$ within the fixed-outline 
The characteristic of LHS lies in that it can make the generated samples distribute as evenly as possible in the parameter space, thus covering a broader solution space and thus increasing the likelihood of finding the global optimum. Then, the stages of global floorplanning and legalization are successively implemented. If a legal solution $\mathbf {Ind}^*$ is obtained after the legalization stage, record this solution as the final solution and terminate the iteration. Otherwise, some randomly selected modules are rotated for $\mathbf{Ind}_i$ and the floorplanning process can be repeated again until the iteration budget $t_{max}=10$ is exhausted.

\begin{algorithm}[!htb]
\caption{The Framework of $CSF$}
\label{alg:fp}
\begin{algorithmic}[1]
\Require Iteration buget $t_{max}$;
\Ensure  A floorplan $\mathbf{Ind}^*=(\mathbf{x}^*,\mathbf{y}^*)$;
\State  Generate a population of floorplan $\mathbf{Pop}=\{\mathbf{Ind}_1,\dots,\mathbf{Ind}_p\}$, where $\mathbf{Ind}_i=(\mathbf{x}^i,\mathbf{y}^i)(i=1,\dots,p)$ is the initial floorplan sampled within the fixed-outline by LHS;
\State Set $ t=1$;
\While {$t<t_{max}$}
\State $\mathbf{Pop}=GFloorplan(f_g,\mathbf{Pop})$; \hfill /*Global floorplanning*/
\State Implement legalization stage to get $\mathbf{Ind}^*$ \hfill /*Legalization*/
\If{$\mathbf{Ind}^*$ is legal}
\State $\mathbf{break}$;
\EndIf
 \State {Rotate randomly selected modules for each individual $\mathbf{ind}_i(i=1,\dots,p)$;}\hfill /*Random rotation of modules*/
 \State t++;
\EndWhile
\end{algorithmic}
\end{algorithm}

\subsection{The Conjugate Subgradient Algorithm Assisted by Q-learning}
The global floorplanning and the legalization address problems \eqref{prob:gf} and \eqref{prob:lg} by CSAQ, which employs a population-based iteration scheme with the principle of CSA. Thanks to the gradient-like definition and the stochastic characteristics of subgradient, the CSA incorporates both the fast convergence of gradient-based algorithms and the global convergence of metaheuristics. The parameter $c$ of CSA (Algorithm \ref{alg:CSA})  plays a crucial role in achieving the balance between exploration and exploitation, which is updated by a Q-learning-assisted strategy~\cite{Li2023g}.




\begin{table}[!htp]
\caption{The Q-table for the update of $c$.}\label{ql:qt}
    \centering
    \begin{tabular}{ccccc}
    \hline\hline
    \multirow{2}{*}{State } & \multicolumn{4}{c}{Action}\\
    \cline{2-5}
       &$a_1$&$a_2$&…&$a_m$\\
    \hline
    $s_1$/$\mathbf{Ind}_1$ & $Q(s_1,a_1)$ & $Q(s_1,a_2)$ & {    …    } & $Q(s_1,a_m)$\\
    $s_2$/$\mathbf{Ind}_2$ & $Q(s_2,a_1)$ & $Q(s_2,a_2)$ & {    …    } & $Q(s_2,a_m)$\\
    $\vdots$& $\vdots$&$\vdots$ &  & $\vdots$\\
    $s_p$/$\mathbf{Ind}_p$ & $Q(s_p,a_1)$ & $Q(s_p,a_2)$ & {    …    } & $Q(s_p,a_m)$\\
    \hline\hline
    \end{tabular}
\end{table}
For a solution population $\mathbf{Pop}$ of size $p$, a $p\times m$ Q-table is constructed as Table \ref{ql:qt}, where each state represents an individual, and an action corresponds to an alternative value of $c$~\cite{Li2023g}. In this paper, the $p$ and $m$ are set to 5. Subsequently, the scaling factor $c$ corresponding to state $s_i$ ($\mathbf{Ind}_i$) is updated via a probability distribution
\begin{equation} \label{ql:pd}
P_i ^j=\frac{Q^k(s_i,a_j)}{\sum_{l=1}^{m}Q^k(s_i,a_l)},\quad j=1,\dots,m,
\end{equation}
where $Q^k(s_i,a_j)$ is the Q-value of state-action pair $(s_i,a_j)$ at the $k$-th iteration. 
At each iteration, the Q-values are updated by \eqref{ql:update} where the reward function $R(s_i,a_j)$ is confirmed by

\begin{equation}\label{reward_f}
    R(s_i,a_j)=(f_{pre}-f_{post})/100,
\end{equation}
where $f_{pre}$ and $f_{post}$ are the pre-action value and the post-action value of the objective function $f$ to be minimized.

With the probability distribution confirmed by \eqref{ql:pd}, the parameter $c$ is updated every $k_t$ generation. Accordingly, we get the framework of CSAQ as presented in Algorithm \ref{alg:CSAQ}.

\begin{algorithm}[!htb]
\caption{$CSAQ(f,\mathbf{Pop})$}\label{alg:CSAQ}
\begin{algorithmic}[1]
\small \Require  The objective $f$, the population $\mathbf{Pop}=\{\mathbf{Ind}_i,i=1,\dots,p\}$; 
\Ensure The updated solution population $\mathbf{Pop}$;
\State Initialize the parameters $k_t$ and the Q-table (Table \ref{ql:qt}) uniformly;
\For{$k=1$ to $k_{max}$}
\For{$i = 1$ to $p$}
    \If{$k = =1$}
    \State Set $R(s_i,a_j)=0$ for all $j\in\{1,\dots,m\}$, initialize $c_j$;
    \State $\mathbf{d}_{i,0}=\mathbf{0}$, $\mathbf{Ind}^0_i=\mathbf{Ind}_i$, $\mathbf g_{i,0}\in\partial f(\mathbf{Ind}_i ^{0})$;
    \EndIf
\If{$k\pmod{k_t} =0$}
\State Update $Q^k(s_i,a_j)$ according to \eqref{ql:update};
\State Update $c_i$ according to the distribution \eqref{ql:pd};
\EndIf
\State Calculate the subgradient $\mathbf g_{i,k}\in\partial f(\mathbf{Ind}_i ^{k-1})$;
\State Calculate the  parameter $\eta _{i,k}=\frac{\mathbf g_{i,k} ^T(\mathbf g_{i,k}-\mathbf g_{i,k-1})}{\lVert \mathbf g_{i,k-1}\rVert _2 ^2}$;
\State Calculate the conjugate direction $\mathbf d_{i,k}=-\mathbf g_{i,k}+\eta _{i,k} \mathbf d_{i,k-1}$;
\State Calculate the step size $\hat a_{i,k}=\frac{c_i}{\lVert \mathbf d_{i,k}\rVert _2}$;
\State Update the solution $\mathbf{Ind}_i ^k=\mathbf{Ind}_i ^{k-1}+\hat a_{i,k} \mathbf d_{i,k}$;
\EndFor
\EndFor
\State Set $\mathbf{Pop}=\{\mathbf{Ind}^{k_{max}}_i,i=1,\dots,p\}$.
\end{algorithmic}
\end{algorithm}

\subsection{The Global Floorplanning}

The global floorplanning is implemented by addressing problem \eqref{prob:gf} using the CSAQ described  by Algorithm \ref{alg:CSAQ}. Because the weight setting of $f_g$ depends on the case to be investigated, we employ a dynamic setting of objective weight $\lambda$. As presented in Algorithm \ref{alg:gf}, the CSAQ is executed iteratively for the parameter setting $\lambda \leftarrow q \lambda$, where $q$ is set as $1.3$ in this paper. In this way, it tries to achieve a promising result striking a balance between wirelength and constraint violation. The iteration process ceases if the total overlapping area of all individuals in the solution population is less than $A/v$, where $A$ is the sum of the areas of all modules and $v$ is a preset constant.


\begin{algorithm}[!htb]
\caption{$GFloorplan(f_g,\mathbf{Pop})$}
\label{alg:gf}
\begin{algorithmic}[1]
 \Require  The objective $f_g$, the population $\mathbf{Pop}=\{\mathbf{Ind}_1,\dots,\mathbf{Ind}_p\}$;
\Ensure  The updated solution population $\mathbf{Pop}^*$;
\State Initialize parameters $\alpha$, $\lambda$ and $\mu$ of the objective $f_g$ (Eqn. \eqref{prob:gf});
\While {the termination condition is not met}
\State $\mathbf{Pop}=CSAQ(f_g,\mathbf{Pop})$;

\State $\lambda=q \lambda$;
\EndWhile
\State $\mathbf{Pop}^*=\{\mathbf{ind}_1,\dots,\mathbf{ind}_p\}$.
\end{algorithmic}
\end{algorithm}

\subsection{The Legalization}\label{sub:LG}
Li \emph{et al.}~\cite{Li2023} proposed a tailored legalization strategy based on the constraint graph (LA-CG), which generates promising results on large-scale floorplanning cases with soft modules. However, our preliminary experiment indicated that it does not work well on small cases without soft modules. Accordingly, we propose an improved legalization algorithm based on constraint graph (ILA-CG). Moreover, we also perform the legalization by solving problem \eqref{prob:lg}. The CSAQ has been demonstrated to be an efficient optimizer, then we get an efficient legalization algorithm based on CSAQ (LA-CSAQ).


\subsubsection{The legalization algorithm based on constraint graph}\label{ssub:LCG}
The LA-CG proposed by Li \emph{et al.}~\cite{Li2023} starts with the construction of the HCG and the VCG, which are then translated into a floorplan without overlap.  If all modules are placed in the constrained region, we get a legal floorplan; otherwise, there are some modules placed outside the fixed outline, and constraint relationships in the critical path~\footnote{The so-called critical path refers to a path composed of blocks that constrain each other in the same direction and are closely arranged~\cite{14}.} of a CG are selected to be adjusted to eliminate the out-of-bound violation. 
Since the process to eliminate horizontal out-of-bound violation is similar to that in the vertical direction, we only present the details of horizontal legalization.

To eliminate the horizontal out-of-bound violation, the LA-CG proposes several definitions for modules and arcs in CG~\footnote{Please refer to Ref. \cite{Li2023} for the rigorous definitions.}.
\begin{itemize}
    \item For module $A$, the maximum adjustment range of the x-coordinate tolerated by the HCG is defined as the \emph{horizontal slack} $S_A^x$; similarly, $S_A^y$ is the \emph{vertical slack} defined as the maximum adjustment range of the y-coordinate.

    \item If  $S_A^x=S_B^x=0$, the arc $A\to B$ (representing a relationship between $A$ and $B$) in HCG is termed as a \emph{horizontal critical relationship} , which means the x-coordinates of modules $A$ and $B$  cannot be adjusted without overlapping with other modules; A \emph{vertical critical relationship} $A\uparrow B$ in the VCG can be defined similarly. The arcs in the critical path of a HCG/VCG are a critical relationship.

    \item For a horizontal critical relationship $A\to B$, The weight is defined by
    \begin{equation} \label{prob:FOFP0}
weight(A\to B)=
\begin{cases}
S^y_A-h_B & \text{if }y_A\leq y_B,\\
S^y_B-h_A & \text{otherwise}.
\end{cases}
\end{equation}
The weight for vertical critical relationship $A\uparrow B$ can be defined similarly.

\item Modules $A$ and $B$ have a \emph{compressible relationship} in the horizontal/vertical direction if $A$ or $B$ may be moved horizontally/vertically such that the floorplan is more compact.
\end{itemize}
Then, the LA-CG modifies the HCG as follows.
\begin{enumerate}
    \item If an arc $A\to B$ is \emph{compressible}, it is removed from the HCG.
    \item The \emph{critical relationship} $A\to B$ with the maximum weight is removed from the HCG, and a corresponding relationship is inserted to the VCG at the same time.
\end{enumerate}
The modification of VCG is performed similarly. By modifying the critical relationship with the maximum weight, the LA-CG aims to reduce the vertical placement range to the greatest extent, employing a greedy strategy that sometimes fail to achieve the optimal modification in terms of wirelength minimization.

Although modification of the critical relationship with the maximum weight can reduce the vertical placement range to the greatest extent, it is a greedy strategy that sometimes does not get the best modification in terms of wirelength minimization. To remedy the deficiency of the LA-CG, we propose an improved legalization strategy in the $x$-direction (named as  $ILG_x$)  presented in Algorithm \ref{alg:lhd}, and the $ILG_y$ for legalization in the $y$-direction is implemented similarly.

\begin{algorithm}[!htb]
\caption{ $ILG_x(\mathbf{W^*},HCG,VCG)$}\label{alg:lhd}
\begin{algorithmic}[1]
\small \Require Width $\mathbf{W^*}$ of the fixed outline, the CG pair $(HCG,VCG)$;
\Ensure  $(HCG,VCG)$ after horizontal legalization.
\State Calculate the width $\mathbf{W_d}$ of the current floorplan and locate all the critical relationships in the $HCG$;
\While{$\mathbf{W_d>W^*}$}
    \If{there exists a \emph{compressible} $A\to B$ }
        \State Remove $A\to B$ from the HCG;
    \Else
    \State Calculate the weights of all critical relationships by \eqref{prob:FOFP0};
\State Select critical relationships $\{A_i\to B_i,i=1,\dots,k\}$ with $k$ greatest weights;
    \State Calculate the selection probability $Pw_i$ for $A_i\to B_i$ ($i=1,\dots,k$);
    \State Set $A\to B$ as a relationship selected from $\{A_i\to B_i,i=1,\dots,k\}$ according to the distribution $\{Pw_i,i=1,\dots,k\}$;
    \State Remove $A\to B$ from the HCG;
    \EndIf
    \If{$y_A\leq y_B$}
    \State Insert $A\uparrow B$ to the VCG;
    \Else
    \State Insert $B\uparrow A$ to the VCG;
    \EndIf
    \State Calculate the width $\mathbf{W_d}$ and locate all critical relationships in the $HCG$;
\EndWhile
\end{algorithmic}
\end{algorithm}

 According to the weight values quantified by \eqref{prob:FOFP0}, the proposed $ILG_x$ gets $k$ candidate relationships by selecting critical relationships with top $k$ weight values. Then, $k$ candidate relationships are sorted in descending order by the potential wirelength increment, and we get a probability distribution $\{Pw_i,i=1,\dots,k\}$, where $Pw_i>Pw_{i+1}$ for all $i\in\{1,\dots,k-1\}$. The distribution plays a significant role in the performance of the CG-based legalization, and we perform a preliminary study in Section \ref{sec:lgcom}.


By alternatively executing $ILG_x$ and $ILG_y$ until a legal floorplan is obtained or the maximum number of iterations $N_{max}$ is reached, we get the improved legalization algorithm based on constraint
graph (ILA-CG) presented in Algorithm \ref{alg:ILGDG}, trying to reduce the wirelength while legalization is implemented. 

\begin{algorithm}[htp]
\caption{$ILA-CG(\mathbf{W, H,Pop})$}\label{alg:ILGDG}
\begin{algorithmic}[1] 
\Require  The population $\mathbf {Pop}=(\mathbf{Ind}_1,\dots,\mathbf{Ind}_p)$, width $\mathbf{W^*}$ and height $\mathbf{H^*}$ of the fixed outline;
\Ensure The post-legalization solution/floorplan $\mathbf {Ind}^*=(\mathbf{x}^*,\mathbf{y}^*)$;
\For{$i=1,\dots,p$}

\For{$k=1$ to $N_{max}$}
\State Generate $HCG_i$ and $VCG_i$ according to $\mathbf {Ind}_i$;
\State Update $\mathbf {Ind}_i$ according to the $HCG_i$ and the $VCG_i$;
\State $(HCG'_i,VCG'_i)=ILG_x(\mathbf{W^*},HCG_i,VCG_i)$;\hfill /*Horizontal legalization*/
\State $(HCG'_i,VCG'_i)=ILG_y(\mathbf{H^*},HCG_i,VCG_i)$;\hfill /*Vertical legalization*/
\State Update $\mathbf {Ind}_i$ according to the $HCG'_i$ and the $VCG'_i$;
\State Calculate the width $\mathbf{W_d}$ and height $\mathbf{H_d}$ of the current floorplan;
     \If{$\mathbf{W_d\le W^*}$ and $\mathbf{H_d\le H^*}$}
         \State $\mathbf {Ind}^*=\mathbf {Ind}_i$;
         \State $\mathbf{return}$;
     \EndIf
\EndFor
\EndFor
\end{algorithmic}
\end{algorithm}



\subsubsection{The legalization algorithm based on CSAQ}
\begin{figure}[!htb]
    \centering
    \begin{subfigure}[b]{0.49\textwidth}
        \centering
        \includegraphics[width=\textwidth]{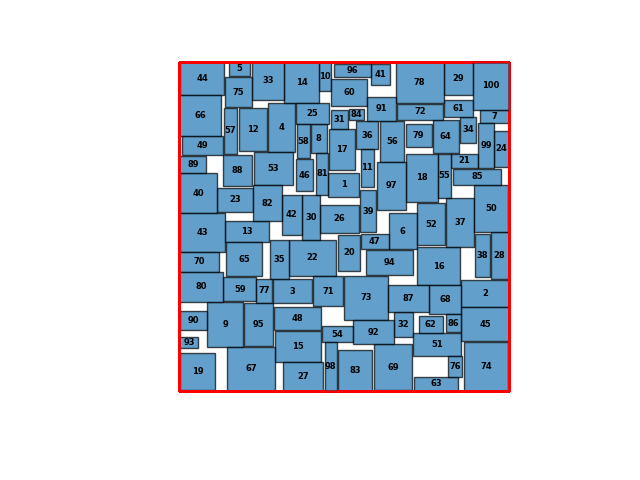}
        \caption{The floorplan before compress.}
        \label{fig:before CG}
    \end{subfigure}
    \begin{subfigure}[b]{0.49\textwidth}
        \centering
        \includegraphics[width=\textwidth]{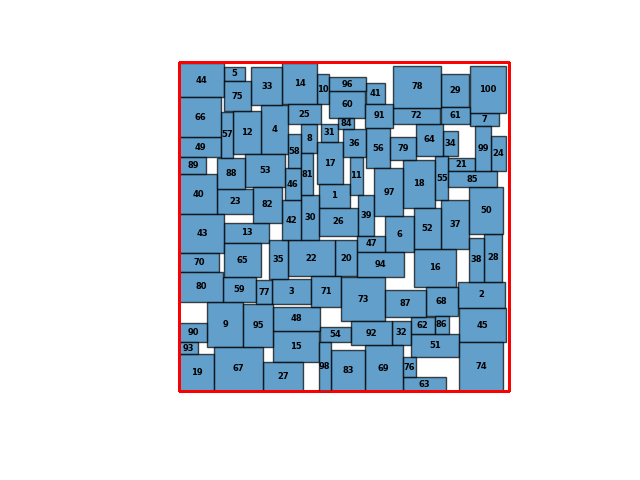}
        \caption{The floorplan after compress.}
        \label{fig:after CG}
    \end{subfigure}
    \caption{The floorplan compression implemented by CG.}
    \label{fig:compress}
\end{figure}

Since the ILA-CG does not explicitly optimize the wirelength, it sometimes generates a legalized floorplan with poor wirelength metric. Then,
Sun \emph{et al.}~\cite{Sun2024} proposed to directly optimize \eqref{prob:lg} by CSA to get a legal floorplan. To further improve the performance of CSA-based legalization, the CSAQ is employed in this paper, 
and we get the legalization algorithm based on CSAQ (LA-CSAQ) presented in Algorithm \ref{alg:lgcsa}. The CSAQ used in the legalization stage is similar to that deployed in the global floorplanning stage, and the iteration process is terminated when either the maximum number of iterations is reached or a legal individual (floorplan) is obtained. Optimization of problem \eqref{prob:lg} by CSAQ sometimes gets an incompact floorplan (Fig. \ref{fig:before CG}), which incurs unnecessary wirelength costs. Accordingly, the CG pair (HCG,VCG) is constructed to update the legalized floorplan to make the modules compactly arranged (Fig. \ref{fig:after CG}).

\begin{algorithm}[!htb]
\caption{$LA-CSAQ(\mathbf{f_l, Pop})$}
\label{alg:lgcsa}
\begin{algorithmic}[1]
\Require  The population $\mathbf {Pop}=(\mathbf{Ind}_1,\dots,\mathbf{Ind}_p)$;
\Ensure The post-legalization solution/floorplan $\mathbf {Ind}^*=(\mathbf{x}^*,\mathbf{y}^*)$;
\State Initialize parameters $\lambda$ and $\mu$ of the objective $f_l$, the parameter $k_t$, initialize the Q-table (Table \ref{ql:qt}) uniformly;
\For{$k=1$ to $k_{max}$}
\For{$j = 1$ to $p$}
    \If{$k = =1$}
    \State Set $R(s_j,a_i)=0$ for all $i\in\{1,\dots,m\}$, initialize $c_j$;
    \State $\mathbf{d}_{j,0}=\mathbf{0}$, $\mathbf{Ind}^0_j=\mathbf{Ind}_j$, $\mathbf g_{j,0}\in\partial f_l(\mathbf{Ind}_j ^{0})$;
    \EndIf
\If{$k\pmod{k_t} =0$}
\State Update $Q^k(s_j,a_i)$ according to \eqref{ql:update};
\State Update $c_j$ according to \eqref{ql:pd};
\EndIf
\State Calculate the subgradient $\mathbf g_{j,k}\in\partial f_l(\mathbf{Ind}_j ^{k-1})$;
\State Calculate the  parameter $\eta _{j,k}=\frac{\mathbf g_{j,k} ^T(\mathbf g_{j,k}-\mathbf g_{j,k-1})}{\lVert \mathbf g_{j,k-1}\rVert _2 ^2}$;
\State Calculate the conjugate direction $\mathbf d_{j,k}=-\mathbf g_{j,k}+\eta _{j,k} \mathbf d_{j,k-1}$;
\State Calculate the step size $\hat a_{j,k}=\frac{c_j}{\lVert \mathbf d_{j,k}\rVert _2}$;
\State Update the solution $\mathbf{Ind}_j ^k=\mathbf{Ind}_j ^{k-1}+\hat a_{j,k} \mathbf d_{j,k}$;
\If{$f_l(\mathbf{Ind}_j ^{k})=0$}
\State $\mathbf {Ind}^*=\mathbf{Ind}_j ^k$;
\State $\mathbf{return}$;
\EndIf
\EndFor
\EndFor
\State Update $\mathbf{Ind}^*$ by the constraint graphs to get tight floorplans.
\end{algorithmic}
\end{algorithm}

\section{Experimental Results}\label{sec:Exp}

To verify the competitiveness of  CSF, we perform numerical experiments on the  GSRC and MCNC benchmarks, where the I/O pads are fixed at the given coordinates for all of the test circuits, and all modules are set as hard modules. The characteristics of the benchmarks are presented in Tab. \ref{tab:benchmark}, and the compared algorithms are implemented in C++ and run on a 64-bit Windows laptop with  2.4 GHz Intel Core i7-13700H and  16-GB memory.

\begin{table}[!htb]
\caption{Characteristics of the Benchmarks.}
\label{tab:benchmark}
\centering
\resizebox{\columnwidth}{!}{
\begin{tabular}{cccccccc}
\hline \hline
\multirow{2}{*}{Benchmark} & \multirow{2}{*}{Instance}  & \multirow{2}{*}{\# Modules} & \multirow{2}{*}{\# IO Pins} & \multirow{2}{*}{\# Pins} & \multirow{2}{*}{\# Nets} & \multicolumn{2}{c}{Die Size ($\mu m^2$)} \\
\cline{7-8}
 & & & & & & $s1$ & $s2$\\
 \hline
 \multirow{2}{*}{MCNC} & ami33 & $33$ & $42$ & $480$ & $123$ & $1153.22\times 1153.22$ & $2058\times 1463$\\
 & ami49 & $49$ & $22$ & $931$ & $408$ & $6384.53\times 6384.53$ & $7672\times 7840$\\
 \hline
 \multirow{3}{*}{GSRC} & n100 & $100$ & $334$ & $1873$ & $885$ & $454.341\times 454.341$ &$800\times 800$\\
 & n200 & $200$ & $564$ & $3599$ & $1585$ & $449.5\times 449.5$ &$800\times 800$\\
 & n300 & $300$ & $569$ & $4358$ & $1893$ & $560.487\times 560.487$ &$800\times 800$\\
\hline \hline
\end{tabular}
}
\end{table}

As presented in Tab. \ref{tab:benchmark}, the fixed-outline floorplanning is performed within two different settings of die size. For setting $s1$, the die sizes are generated by \eqref{boundry} with $R=1$ and $\gamma=15\%$. The size setting  $s2$ is cited from Ref. \cite{2025Per-RMAP} to validate the wirelength minimization capability of the CSF. By numerical experiments, we first show how the incorporated Q-learning enhances the performance of CSF, and then, compare it with several selected floorplanning algorithms. The comparison is performed for the cases without  soft modules to highlight how the CSF efficiently addresses the great challenge of minimizing the wirelength within a fixed outline for the scenarios containing only hard modules.

 In the proposed CSAQ, the Q-learning strategy is introduced to automatically regulate the scale factor $c$ of the CSA. Influenced by module sizes and the primary objectives of different stages, we design distinct action groups for updating the $c$ across various benchmarks. Specifically:

 \begin{itemize}
     \item For the global floorplanning stage:
         \begin{itemize}
          \item On GSRC benchmarks, the action groups are [8, 12, 15, 20, 25];
          \item For \emph{ami33}, the action groups are [80, 100, 120, 140, 160];
          \item For \emph{ami49}, the action groups are [130, 180, 220, 270, 330].
          \end{itemize}
     \item For the legalization stage:
         \begin{itemize}
          \item On GSRC benchmarks, the action groups are [0.1, 0.8, 5, 10, 20];
          \item For \emph{ami33}, the action groups are [1, 8, 30, 60, 90];
          \item For \emph{ami49}, the action groups are [10, 50, 100, 150, 200].
         \end{itemize}
 \end{itemize}

\subsection{The Positive Effects of Q-learning}\label{sec:gfcom}


The positive effects of Q-learning are validated by comparing the performance of different  global floorplanning (GP) and legalization (LA) strategies, where the $\alpha_0,\gamma_0,m_0$ of \eqref{ql:update} and \eqref{ql:Qm} are 0.4, 0.8, 0.6 respectively, and the weights $\alpha,\lambda,\mu$ of \eqref{prob:gf} and \eqref{prob:lg} as well as the parameters $k_t,k_{max},c_0$ of CSA and CSAQ are presented in Tab. \ref{tab:parameter1}. The fixed-outline die sizes are set as the $s1$ of Tab. \ref{tab:benchmark}.

\begin{table}[!htb]
\caption{Parameter settings of CSA and CSAQ in different stage.}
\label{tab:parameter1}
\centering
\resizebox{\columnwidth}{!}{
\begin{tabular}{cccccccccc}
\hline \hline
Stage & Optimizer  & Benchmark & $k_t$ & $k_{max}$ & $\alpha$ & $\lambda$ & $\mu$ & $c_0$ & $v$ \\
\hline
\multirow{4}{*}{GP} & \multirow{2}{*}{CSA} & MCNC & - & 50 & 1 & 20 & 10 & 1000 & 10 \\
\cline{3-10}
 & & GSRC & - & 50 & 1 & 20 & 100 & 100 & 100 \\
\cline{2-10}
 & \multirow{2}{*}{CSAQ} & MCNC & 10 & 50 & 1 & 20 & 10 & 330 & 10\\
\cline{3-10}
 & & GSRC & 40 & 200 & 1 & 20 & 100 & 25 & 100\\
 \hline
\multirow{4}{*}{LA} & \multirow{2}{*}{CSA} & MCNC & - & 2000 & - & 1 & 10 & 500 & -\\
\cline{3-10}
 & & GSRC & - & 2000 & - & 1 & 10 & 50 & -\\
\cline{2-10}
 & \multirow{2}{*}{CSAQ} & MCNC & 100 & 2000 & - & 1 & 10 & 100 & -\\
\cline{3-10}
 & & GSRC & 100 & 5000 & - & 1 & 10 & 10 & -\\
\hline \hline
\end{tabular}
}
\end{table}



\subsubsection{The performance of CSAQ-based global floorplanning}\label{ssub:Qglobal}

\begin{table}[!hbp]
\caption{Performance comparison between CSA- and CSAQ-based global floorplanning.}
\label{tab:my_labe2}
\centering
\resizebox{\columnwidth}{!}{
\begin{tabular}{cccccccccc}
\hline \hline
\multirow{2}{*}{Circuit} & \multirow{2}{*}{Method}&\multicolumn{3}{c}{Runtime} & \multirow{2}{*}{SR($\%$)} & \multirow{2}{*}{HPWL($\mu m$)} & \multirow{2}{*}{IR($\%$)} & \multirow{2}{*}{ p-value} &\multirow{2}{*}{R} \\
\cline{3-5}
 &  & $t_{g}$($s$) & $t_{l}$($s$) & $t_w$($s$) & &  & &   \\
 \hline

\multirow{2}{*}{ami33} & CSF-cc  & 0.008 & 0.006 & \textbf{0.23} & 73.3 & 133702 & \multirow{2}{*}{1.70} & \multirow{2}{*}{0.09} & \multirow{2}{*}{$\sim$} \\
\cline{2-7}
 & CSF-qc  & 0.031 & 0.005 & 0.66 & \textbf{80}& \textbf{131427} & & \\
\hline
\multirow{2}{*}{ami49} & CSF-cc  & 0.019 & 0.018 & \textbf{0.31} & 96.6 & \textbf{916042} & \multirow{2}{*}{-0.89} & \multirow{2}{*}{0.11} & \multirow{2}{*}{$\sim$} \\
\cline{2-7}
 & CSF-qc  & 0.11 & 0.016 & 0.45 & \textbf{100}& 924246 & & \\
 \hline
\multirow{2}{*}{n100} & CSF-cc  & 0.22 & 0.07 & \textbf{0.35} & 83.3 & 293202 & \multirow{2}{*}{0.14} & \multirow{2}{*}{$0.24$} & \multirow{2}{*}{$\sim$} \\
\cline{2-7}
 & CSF-qc  & 1.35 & 0.06 & 1.52 & \textbf{100} & \textbf{292780} & & \\
 \hline
\multirow{2}{*}{n200} & CSF-cc  & 0.63 & 0.16 & \textbf{0.82} & 96.6 & 524905 & \multirow{2}{*}{0.32} & \multirow{2}{*}{$0.03$} & \multirow{2}{*}{+} \\
\cline{2-7}
 & CSF-qc  & 4.27 & 0.13 & 4.62 & \textbf{100} & \textbf{523229} & & \\
 \hline
\multirow{2}{*}{n300} & CSF-cc  & 1.47 & 0.34 & \textbf{1.98} & 90 & 595735 & \multirow{2}{*}{0.33} & \multirow{2}{*}{$0.28$} & \multirow{2}{*}{$\sim$} \\
\cline{2-7}
 & CSF-qc  & 10.89 & 0.21 & 11.05 & \textbf{100} & \textbf{593784} & & \\
\hline \hline
\end{tabular}
}
\end{table}

To validate the promising effect of CSAQ on the global floorplanning stage, we compare two variants of CSF, 1) the CSF-cc implementing both global floorplanning and legalization by CSA and 2) the CSF-qc implementing global floorplanning and legalization  by CSAQ and CSA, respectively. 
Due to the random characteristics of CSA, the performances are compared by averaged results of $30$ independent runs. The average runtime of a single run of global floorplanning ($t_{g}$), the average runtime of a single run of legalization ($t_{l}$), the average runtime to complete the entire floorplanning ($t_w$), the success rate (SR), the average wirelength (HPWL) and the improvement ratio(IR) of the HPWL are included in Table \ref{tab:my_labe2}. Moreover, the one-sided Wilcoxon rank-sum test is performed based on the wirelength data of 30 independent runs, and the p-values and the test results (R) with a significance level of $0.05$ are also demonstrated in Table \ref{tab:my_labe2}, where ``+'' indicates that CSF-qc significantly outperforms CSF-cc and ``$\sim$'' means that the difference is not significant.

Compared to CSF-cc, CSF-qc demonstrates an enhanced success rate  at the cost of a marginally longer \(t_g\). However, the runtime of leglaization \(t_l\) is shorter, which indicates that the better floorplanning results obtained by CSAQ can even shorten the runtime of legalization implemented by CSA. Consequently, the CSF-qc is competitive to CSF-cc in the terms of averaged wirelength for most of the test benchmarks.
However, the  CSAQ-assisted GP alone cannot improve the floorplanning results to a large extent, and the hypothesis test shows that CSF-qc only outperforms CSF-cc significantly on the \emph{n200} case.

\subsubsection{The superiority of CSAQ-based legalization to CSA-based legalization}\label{ssub:Qlg}
The CSF-qc is further compared with the CSF-qq, where both global floorplanning and legalization are implemented by CSAQ. The results are presented in Table \ref{tab:my_labe3}.


\begin{table}[!htp]
\caption{Performance comparison between CSA- and QCSA-based legalization.}
\label{tab:my_labe3}
\centering
\resizebox{\columnwidth}{!}{
\begin{tabular}{cccccccccc}
\hline \hline
\multirow{2}{*}{Circuit} & \multirow{2}{*}{Method}&\multicolumn{3}{c}{Runtime} & \multirow{2}{*}{SR($\%$)} & \multirow{2}{*}{HPWL($\mu m$)} & \multirow{2}{*}{IR($\%$)} & \multirow{2}{*}{p-value} &\multirow{2}{*}{R} \\
\cline{3-5}
 &  & $t_{g}$($s$) & $t_{l}$($s$) & $t_w$($s$) & &  & &   \\
 \hline
 \multirow{2}{*}{ami33} & CSF-qc & 0.031 & 0.007 & \textbf{0.66} & 80 & 131427 & \multirow{2}{*}{36.52} & \multirow{2}{*}{$4.79 \times 10^{-11}$} & \multirow{2}{*}{+} \\
\cline{2-7}
 & CSF-qq  & 0.032 & 0.11 & 0.91 & \textbf{83.3} & \textbf{83427} & & \\
\hline
\multirow{2}{*}{ami49} & CSF-qc & 0.11 & 0.017 & \textbf{0.45} & 100 & 924246 & \multirow{2}{*}{2.02} & \multirow{2}{*}{0.21} & \multirow{2}{*}{\~} \\
\cline{2-7}
 & CSF-qq  & 0.10 & 0.13 & 0.58 & 100 & \textbf{905566} & & \\
\hline
\multirow{2}{*}{n100} & CSF-qc & 1.35 & 0.06 & \textbf{1.52} & 100 & 292880 & \multirow{2}{*}{0.93} & \multirow{2}{*}{$7.92 \times 10^{-5}$} & \multirow{2}{*}{+} \\
\cline{2-7}
 & CSF-qq  & 1.33 & 0.25 & 1.76 & 100& \textbf{290156} & & \\
 \hline
\multirow{2}{*}{n200} & CSF-qc  & 4.27 & 0.18 & \textbf{4.62} & 100& 523229 & \multirow{2}{*}{0.77} & \multirow{2}{*}{$9.30 \times 10^{-7}$} & \multirow{2}{*}{+} \\
\cline{2-7}
 & CSF-qq  & 4.28 & 0.64 & 5.35 & 100& \textbf{519211} & & \\
 \hline
\multirow{2}{*}{n300} & CSF-qc & 10.89 & 0.16 & \textbf{11.05} & 100 & 593784 & \multirow{2}{*}{0.81} & \multirow{2}{*}{$4.76 \times 10^{-4}$} & \multirow{2}{*}{+} \\
\cline{2-7}
 & CSF-qq  & 11.21 & 0.92 & 12.40 & 100& \textbf{588951} & & \\
\hline
\hline
\end{tabular}
}
\end{table}

It is demonstrated that by employing the Q-learning strategy, the CSAQ leads to significantly better legalization results at the expense of slightly increased runtime. The statistical test indicates that CSF-qq outperforms CSF-qc on four benchmark circuits. Although the test does not demonstrate significant superiority of CSF-qq in the \emph{ami49} case, it still beats CSF-qc in terms of the average wirelength.


\subsection{Comparison between CG- and QCSA-based legalization algorithms}\label{sec:lgcom}
To further confirm the superiority of the CSAQ-based legalization algorithm, we compare the LA-CSAQ (Algorithm \ref{alg:lgcsa}) with the LA-CG ~\cite{Li2023} by starting the legalization with initial floorplans generated by the CSAQ based global floorplanning algorithm $GFloorplan(f_g,\mathbf{Pop})$(Algorithm \ref{alg:gf}). Meanwhile, two variants of the ILA-CG (Algorithm \ref{alg:ILGDG}) are investigated to highlight the superiority of LA-CSAQ.
Both variants of the ILA-CG set $k=3$ with two different probability distributions.
\begin{itemize}
    \item ILA-CGm: $Pw_1 = 0.9$, $Pw_2 = 0.05$, $Pw_3 = 0.05$;
    \item ILA-CGs: $Pw_1 = 1$, $Pw_2 = 0$, $Pw_3 = 0$.
\end{itemize}

The statistical results of $30$ independent runs are included in Table \ref{tab:legcom}, where  MinW and HWSD represent the wirelength of minimum value and the standard deviation, respectively. 

\begin{table}[!htp]
\caption{Performance Comparison Between CG and QCSA-based Legalization Algorithms}
\label{tab:legcom}
\centering
\resizebox{\columnwidth}{!}{
\begin{tabular}{ccccccc}
\hline\hline
Circuit & Method & HPWL($\mu m$) & MinW($\mu m$) & CPU($s$) & SR($\%$) & HWSD($\mu m$)  \\
\hline
\multirow{4}{*}{ami33}
 & LA-CG \cite{Li2023} & 119625 & 88457 & 2.35 & 43.3 & 10342.88 \\
 \cline{2-7}
 & ILA-CGm & 109235 & 81047 & 1.26 & 80 & 104487.27 \\
  \cline{2-7}
 & ILA-CGs & 103136 & 79173 & 2.45 & 30 & 16993.17 \\ \cline{2-7}
 & LA-CSAQ & \textbf{83427} & \textbf{75783} & \textbf{0.91} & \textbf{83.3} & \textbf{5383.99} \\
\hline
\multirow{4}{*}{ami49}
 & LA-CG \cite{Li2023} & 1626100 & 1275200 & 1.03 & 100 & 147349.55 \\
 \cline{2-7}
 & ILA-CGm & 1481960 & 1064360 & 1.23 & 100 & 164380.71 \\
  \cline{2-7}
 & ILA-CGs & 1422600 & 928432 & 2.73 & 100 & 194858.36 \\ \cline{2-7}
 & LA-CSAQ & \textbf{905566} & \textbf{806937} & \textbf{0.58} & 100 & \textbf{62604.82}   \\

\hline
\multirow{4}{*}{n100}
 & LA-CG \cite{Li2023} & 389435 & 369092 & 37.58 & 60 & 18157.76 \\
 \cline{2-7}
 & ILA-CGm & 363257 & 323760 & 31.87 & 53.3 & 14945.68 \\
  \cline{2-7}
 & ILA-CGs & 359603 & 312565 & 35.37 & 60 & 15775.31 \\ \cline{2-7}
  & LA-CSAQ & \textbf{290156} & \textbf{286277} & \textbf{1.76} & \textbf{100} & \textbf{2273.33} \\

\hline
\multirow{4}{*}{n200}
 & LA-CG \cite{Li2023} & 718581 & 698852 & 187.32 & 10 & 24538.32 \\
 \cline{2-7}
 & ILA-CGm & 708955 & 682711 & 163.25 & 30 & 21937.34 \\
  \cline{2-7}
 & ILA-CGs & 699861 & 666166 & 12.67 & 50 & 19825.83 \\ \cline{2-7}
 & LA-CSAQ & \textbf{519211} & \textbf{516038} & \textbf{5.35} & \textbf{100} & \textbf{2391.86} \\

\hline
\multirow{4}{*}{n300}
 & LA-CG \cite{Li2023} & - & - & - & 0 & - \\
 \cline{2-7}
 & ILA-CGm & - & - & - & 0 & - \\
  \cline{2-7}
 & ILA-CGs & - & - & - & 0 & - \\ \cline{2-7}
 & LA-CSAQ & \textbf{588951} & \textbf{580712} & \textbf{12.40} & \textbf{100} & \textbf{4667.06} \\
\hline\hline
\end{tabular}
}
\end{table}

The results show that the improved CG-based legalization algorithms (ILA-CGm and ILA-CGs) generally outperform the LA-CG in terms of both the average wirelength and the minimum wirelength. Since ILA-CGs always selects the candidate correction relationship contributing to the smallest wirelength, it optimizes the wirelength better than ILA-CGm. Compared with the CG-based legalization methods including LA-CG, ILA-CGm and ILA-CGs, the LA-CSAQ can get legal floorplans with the shorter wirelength, demonstrating its superiority in optimization of wirelength.


From the perspectives of the average time and the success rate, the LA-CSAQ gets legal floorplans with the shortest runtime and the highest success rate, and the significantly smaller HWSD values demonstrate its superior robustness regarding the wirelength of floorplan. Although both ILA-CGm and ILA-CGs outperform LA-CG on circuits \emph{ ami33, ami49, n100} and \emph{n200}, none of the CG-based legalization algorithms can eliminate the constraint violations for the \emph{n300} circuit, which further confirms the superior performance of LA-CSAQ in the task of floorplan legalization.




\subsection{Comparison between CSF and the state-of-the-art floorplanning algorithms}
The competitiveness of the proposed CSAQ-based floorplanning algorithm is further validated by comparing the CSF-qq (where both global floorplanning and legalization are addressed by CSAQ) with some selected floorplanning algorithms, the Parquet-4.5~\cite{Saurabh2003}, the Fast-FA~\cite{2006Fast-FA}, the FFA-CD~\cite{Sun2024} and the Per-RMAP~\cite{2025Per-RMAP}. Among them, Parquet-4.5 and Fast-FA are based on the B*-tree coding structure of floorplan, and the optimization of wirelength is addressed by the simulated annealing algorithm; FFA-CD utilizes the distribution evolution algorithm based on the probability model to optimize the orientation of modules and employs CSA to optimize the module coordinates; Per-RMAP models the floorplanning problem as a feasibility-seeking problem, and then deals with the floorplanning problem by introducing a perturbed resettable strategy into the method of alternating projection.

For the fixed-outline floorplanning scenario, the introduction of soft module is beneficial to achieve legal floorplans with short wirelength. However, this paper is dedicated to the scenario of fixed-outline floorplanning without soft modules. Note that the experimental data of Fast-FA and Per-RMAP are taken from Ref. \cite{2025Per-RMAP}, in which the success rates of the benchmark circuits are not presented.

\subsubsection{Floorplanning within compact layout regions}
\begin{figure*}[!htb]
    \centering
    \begin{subfigure}[b]{0.3\textwidth}
        \centering
        \includegraphics[width=\textwidth]{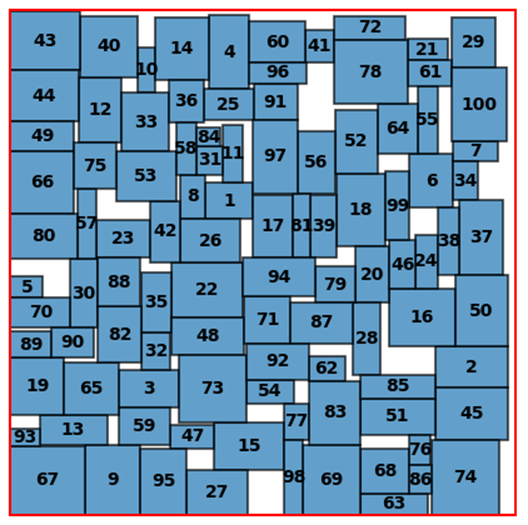}
        \caption{\emph{n100}}
        \label{fig:n100}
    \end{subfigure}
    \hfill
    \begin{subfigure}[b]{0.3\textwidth}
        \centering
        \includegraphics[width=\textwidth]{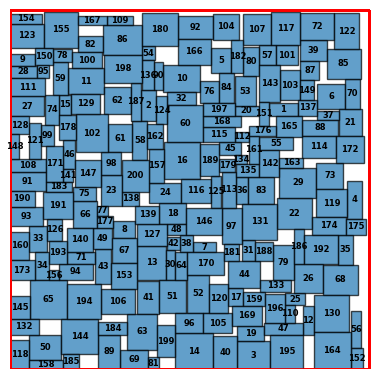}
        \caption{\emph{n200}}
        \label{fig:n200}
    \end{subfigure}
    \hfill
    \begin{subfigure}[b]{0.3\textwidth}
        \centering
        \includegraphics[width=\textwidth]{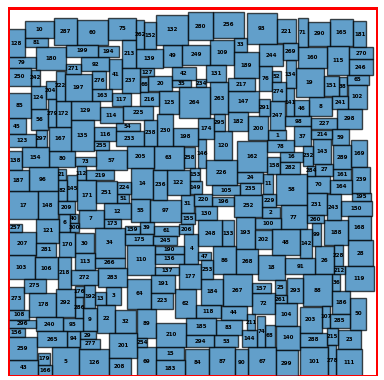}
        \caption{\emph{n300}}
        \label{fig:n300}
    \end{subfigure}
    \caption{The Floorplans of GSRC benechmarks generated by CSF-qq. }
    \label{fig:all}
\end{figure*}

We first validate the performance of  CSF-qq on floorplanning scenarios with a tightly arranged layout region by comparing it with Parquet-4.5 and FFA-CD. The layout region is set as setting $s1$ presented in Tab. \ref{tab:benchmark}.  The generated floorplans of CSF-qq on the GSRC benchmarks are illustrated in Fig. \ref{fig:all}, and the statistical results of 30 independent runs are presented in Table \ref{tab:my_labe5}, where IR represents the wirelength improvement ratio of CSF-qq compared to other methods.

\begin{table}[!htb]
\caption{Performance Comparison for the Floorplanning Restricted in Layout Regions with 15\% Whitespace }
\label{tab:my_labe5}
\centering
\resizebox{\columnwidth}{!}{
\begin{tabular}{cccccc}
\hline\hline
Circuit & Method & HPWL($\mu m$) & CPU($s$) & SR($\%$) & IR($\%$)\\
\hline
\multirow{3}{*}{ami33} & Parquet-4.5 & \textbf{78650} & \textbf{0.16} & 60 & -6.07 \\
\cline{2-6}
 & FFA-CD & - & - & - & - \\
\cline{2-6}
 & CSF-qq & 83427 & 0.91 & \textbf{83.3} & - \\
\hline
\multirow{3}{*}{ami49} & Parquet-4.5 & 962654 & \textbf{0.37} & 60 & 5.93 \\
\cline{2-6}
 & FFA-CD & - & - & - & - \\
 \cline{2-6}
 & CSF-qq & \textbf{905566} & 0.58 & \textbf{100} & - \\
\hline
\multirow{3}{*}{n100} & Parquet-4.5 & 334719 & 1.73 & 100 & 13.31 \\
\cline{2-6}
 & FFA-CD & 293578 & \textbf{0.70} & 100 & 1.17 \\
 \cline{2-6}
 & CSF-qq & \textbf{290156} & 1.76 & 100 & - \\
\hline
\multirow{3}{*}{n200} & Parquet-4.5 & 620097 & 7.78 & 100 & 16.27 \\
\cline{2-6}
 & FFA-CD & 521140 & \textbf{1.95} & 100 & 0.37 \\
 \cline{2-6}
 & CSF-qq & \textbf{519211} & 5.35 & 100 & - \\
 \hline
\multirow{3}{*}{n300} & Parquet-4.5 & 768747 & 16.86 & 100 & 23.39 \\
\cline{2-6}
 & FFA-CD & \textbf{588118} & \textbf{3.44} & 100 & -0.14 \\
 \cline{2-6}
 & CSF-qq & 588951 & 12.40 & 100 & - \\
\hline
\end{tabular}}
\end{table}

The numerical results show that CSF-qq is generally competitive with the compared algorithms. Compared to Parquet-4.5, the wirelength on GSRC benchmarks is improved by $13.31\sim 23.39\%$, the running time is reduced to a large extent. In comparison with FFA-CD, the CSF-qq generates floorplans on \emph{n100} and \emph{n200} with shorter wirelength, and the wirelength of problem \emph{n300} is a bit worse than FFA-CD by -0.14\%. Since a population-based CSA framework assisted by Q-learning is employed by CSF-qq, its running time on GSRC benchmarks is slightly longer than FFA-CD.

Meanwhile, the global convergence of CSF-qq is guaranteed by the optimizer CSAQ. It can address the MCNC benchmarks better, achieving 100\% success rate (SR) on \emph{ami33} and \emph{ami49}. In comparison, the SR of Parquet-4.5 is 60\%, and there is no reference data available for FFA-CD in Ref.\cite{Sun2024}.

\subsubsection{Wirelength minimization within spacious layout regions}

To validate the performance of CSF-qq on minimization of wirelength, we compare it with two representative algorithms, the Fast-SA and the Per-RMAP, where the fixed-outlines of layout regions are generated according to setting $s2$ presented in Tab. \ref{tab:benchmark}~\cite{2025Per-RMAP}. As presented in Tab. \ref{tab:my_labe6}, CSF-qq outperforms both Fast-SA and Per-RMAP on the GSRC benchmarks in terms of wirelength, however, it gets slightly longer wirelength when addressing the MCNC benchmarks. With regard to the running time, both CSF-qq and Per-RMAP beat Fast-SA markedly. Note that the results of Per-RMAP are cited from Ref. \cite{2025Per-RMAP} where it is implemented on a server equipped with 2-way Intel Xeon
Gold 6248R@3.0-GHz CPUs and 768-GB DDR4-2666-MHz
memory, CSF-qq is also competitive to Per-RMAP in terms of running time.


\begin{table}[!htb]
\caption{Performance Comparison on \cite{2025Per-RMAP} Fixed-outline Floorplanning}
\label{tab:my_labe6}
\centering
\resizebox{\columnwidth}{!}{
\begin{tabular}{cccccc}
\hline\hline
Circuit & Method & HPWL($\mu m$) & CPU($s$) & SR($\%$) & IR($\%$)\\
\hline
\multirow{3}{*}{ami33} & Fast-SA & - & - & - & -\\
\cline{2-6}
& Per-RMAP & \textbf{63079} & 0.14 & - & -2.82 \\
 \cline{2-6}
 & CSF-qq & 64857 & 0.14 & 100 & - \\
\hline
\multirow{3}{*}{ami49} & Fast-SA & - & - & - & -\\
\cline{2-6}
& Per-RMAP & \textbf{689296} & 1.70 & - & -5.56 \\
 \cline{2-6}
 & CSF-qq & 727610 & \textbf{0.27} & 100 & - \\
\hline
\multirow{3}{*}{n100} & Fast-SA & 287646 & 10.72 & - & 3.22 \\
 \cline{2-6}
  & Per-RMAP & 282596 & \textbf{1.01} & - & 1.49 \\
 \cline{2-6}
 & CSF-qq & \textbf{278374} & 1.74 & 100  & -\\
\hline
\multirow{3}{*}{n200} & Fast-SA & 516057 & 69.86 & - & 2.29 \\
  \cline{2-6}
 & Per-RMAP & 518722 & \textbf{2.93} & - & 2.79 \\
 \cline{2-6}
 & CSF-qq & \textbf{504258} & 3.75 & 100 & - \\
 \hline
\multirow{3}{*}{n300} & Fast-SA & 603811 & 133.63 & - & 3.05\\
\cline{2-6}
 & Per-RMAP & 626061 & \textbf{4.11} & - & 6.49 \\
 \cline{2-6}
 & CSF-qq & \textbf{585419} & 5.99 & 100 & - \\
\hline
\end{tabular}}
\end{table}

\section{Conclusion}\label{sec:Con}
The state-of-the-art researches demonstrate that the analytic floorplanning algorithms generally outperform the counterparts based on the working principle of combinatorial optimization. These analytic floorplanning algorithms typically employ some gradient-based optimization algorithms addressing the smoothed objectives, which introduces inevitable distortion to the characteristics of the mathematical model. Consequently, the generated floorplan sometimes is not as optimal as expected.

In this paper, we propose to address the original nonsmooth optimization model of floorplanning by the CSA, and the Q-learning strategy is introduced to improve the convergence performance of the CSA-based floorplanning algorithm. Experimental results show that the proposed CSF accelerated by Q-learning (CSF-qq) is competitive to the state-of-the-art fixed-outline floorplanning algorithms in the case that only hard modules are considered. However, the population-based iteration scheme leads to a slight increase runtime in some investigated scenarios, and its performance on small-scale MCNC benchmarks is not as good as that on GSRC problems. Accordingly, our future work will focus on the development of orientation optimization of modules, try to decrease the time complexity of the Q-learning accelerated algorithm, and extend it to the case of large-scale floorplanning with both soft and hard modules.

\bibliographystyle{elsarticle-num}
\bibliography{references}






\end{document}